# Considering users' behaviours in improving the responses of an information base

Babajide Afolabi and Odile Thiery
*Laboratoire Lorrain de Recherche en Informatique et ses Applications (LORIA) Campus Scientifique BP 239, 54506 Vandoeuvre-lès-Nancy, France.*

***ABSTRACT***: *In this paper, our aim is to propose a model that helps in the efficient use of an information system by users, within the organization represented by the IS, in order to resolve their decisional problems. In other words we want to aid the user within an organization in obtaining the information that corresponds to his needs (informational needs that result from his decisional problems). This type of information system is what we refer to as economic intelligence system because of its support for economic intelligence processes of the organisation. Our assumption is that every EI process begins with the identification of the decisional problem which is translated into an informational need. This need is then translated into one or many information search problems (ISP). We also assumed that an ISP is expressed in terms of the user's expectations and that these expectations determine the activities or the behaviors of the user, when he/she uses an IS. The model we are proposing is used for the conception of the IS so that the process of retrieving of solution(s) or the responses given by the system to an ISP is based on these behaviours and correspond to the needs of the user.*

## 1 INTRODUCTION

The overall research interests of our research group centre on modeling and developing Economic Intelligence[1] (EI) systems. We chose to consider Economic Intelligence as "*the process of collection, processing and diffusion of information that has as an objective, the reduction of uncertainty in the making of all strategic decisions*" [1]. Our real interest is in the aid that can be got from using this system to resolve decisional problems. The processes of collection, processing and diffusion can be fully or semi automated, we imagine this automation being powered by an information system (IS) base. The information system that powers a EI belongs to a class of IS referred to as strategic information systems (SIS), strategic in the sense that it contains information that is considered strategic because this information is used in the decisional processes of the organization (and not the type used to run their day to day activities). For example, an IS that permits the decision maker to observe sales by country for a number of years or that permits an information watcher to point up the choices made during the analysis of the result obtained from an information search on the web. This is denoted as S-IS [2], [3]. This type of IS for EI is what we refer to as EIS

It is with this background that we propose a model that will help a user of an EIS in resolving his information search problem (ISP) based on his behaviour (activities) while using such a system.

## 2 RESEARCH CONTEXT

The decisions taken, using an IS, are based on the information found in the IS and are also based on the user that has as an objective, the appropriation of such system for a decision making process. To us, an Economic Intelligence System EIS is a system that combines strategic information systems and user modelling domains. The final goal of an EIS is to help the user or the decision maker in his decision making process.

In the architecture of an EIS as proposed by the research team "SITE"[2] (Modelling and Developing Economic Intelligence Systems of the Lorraine Laboratory of IT Research and its Applications (LORIA) Nancy, France) one can easily identify the following four stages:

1. **Selection:** selection which permits the constitution of the IS of the organisation that can

---

[1] Economic Intelligence is the French equivalent of Competitive Intelligence in some American litteraures and Business Intelligence in some European litteratures.
[2] See http://site.loria.fr

be (i) the production database (used in the daily runnings of the organisation), (ii) all the information support for an information retrieval system (in documentation for an example) or (iii) a SIS based on a data warehouse. This information system is constituted from heterogeneous data and sources with the aid of a filter;
2. **Mapping:** mapping permits all users an access to the data/information in the IS. We are permitting two methods of access to the user: access by exploration and access by request. The exploration is based on a system of hypertexts. The requests are expressed with the aid of Boolean operators. The result of the mapping is a set of information;
3. **Analysis:** in order to add value to the information found, processing techniques are applied to the results, depending eventual requirements. For instance, the assistant of a head of department that we consider as the information watcher can present a summary of the results obtained on the information requested to his head of department;
4. **Interpretation:** this means in general, the possibility of the user of the system being able to make the right decisions. It does not imply that the sole user of the system is the decision maker; other users can include the information watcher. One can see then the interest in capturing the profile of the user in a metadata stored on an information base which can be used to build a specific data mart for a group of decision makers or, even better, a particular user.

Also in this EI process one can identify three main actors:

1. **Decision maker:** this is the individual in the organization that is capable of identifying and posing a problem to be solved in terms of stake, risk or threat that weighs on the organization. In other words, he knows the needs of the organization, the stakes, the eventual risks and the threats the organization can be subjected to.
2. **Information watcher:** this refers to the person within the organization that specializes in the methods of collection and analysis of information. His objective is to obtain indicators (using information) or value added information that the decision makers depend on for his decision process. After receiving the problem to be solved as expressed by the decision maker, the information watcher must translate it into information attributes to be collected and which are used to calculate the indicators.
3. **End user:** this is the final user of the system; it can be either of the previously outlined users or neither of the two. This user is defined depending on which layer of the Economic intelligence system he interacts with.

## 3 THE INFORMATION SYSTEM BASE OF EIS

As highlighted earlier, the success of an EIS system depends on the information system support on which it relies. There are four main stages involved with an EIS: selection, mapping (query formulation), analysis (determination of relevance, etc.), and interpretation. Our interests in this paper lie with the second and third stages.

These two stages suggest a central role for techniques to acquire knowledge about the user and to use that knowledge to guide the mapping and information relevance determination components of the system. This is based on *user modelling*. User modelling implies *adaptivity*, in that the EIS adapts itself to suit the specific user according to knowledge it has acquired about that user. It is also necessary that the system have access to the information content in order to match the user's queries.

### 3.1    User modelling and adaptivity in EIS

*User modelling* can be *defined* as the process of acquiring knowledge about a user in order to provide services or information adapted to their specific requirements [4] and [5]. For approaches to user modelling in information retrieval systems see [6] and [7].

An *adaptive system* is a system that changes its functionality or interface in order to accommodate the differing needs of the users over time [8].

In an EIS, user modelling and *adaptivity* are needed to support the following: (1) the adaptation of user's query by the system in order to meet that user's specific needs as identified by the user model; and (2) the determination of the suitability of information retrieved that is not only *based* on the content of the information, but also on relevance of the content to the user's objective.

### 3.2   Information retrieval, ISP and relevance in economic intelligence

Depending on the author, EI (Competitive Intelligence as it is referred to in certain literature) is a process comprising of a number of steps ([1], [9], and [10]). All of these steps can be summarised into four distinct steps: determination of strategic information need; the collection of the information; interpretation of the information; and diffusion or sharing of the intelligence obtained from the information. These show the importance of having the right (relevant) information, and having it in time for the process of decision making. In order to respond to this question on the relevancy of information, a number of works exists (e.g. [11], [12], etc.). The interactive aspect of information retrieval systems have also been a major object of study as evidenced by [13], [14] and [15]. In order to respond to these exigencies of EI, an ISP has to first be translated into an information need before the relevant information is retrieved [11], [12]. The model we are proposing tries to respond to this aspect of EI by trying to determine the real needs of the user through the studying of his behaviours and his working environment and combining this with his definition of his own needs.

## 4   OUR MODEL FOR THE RESOLUTION OF AN ISP IN EIS

We believe that the resolution of an ISP, in an economic intelligence context, depends on: the understanding of the problem to be resolved; the formulation of the query; the activities that will lead to finding the relevant information; and the means to carry out the search for information (this may include the system available, the totality of information available on which the search is to be conducted, etc). Various models exist in our research team that addresses the different stages of EI. Of these models, one is based on the explicitation of the decisional problem to be resolved by studying its context and insisting on its clarification by the person who first sighted or announced the problem, another is based on the need for good transmission of the problem from the person who sighted the problem to the person who is charged with looking for the information required, etc. (see [3], [16], [17] and [18] for examples). As in the case of decisional problem, for an informational problem to be resolved, the informational need must be clearly identified. This identification constitutes a major set back as the user may not be able to express his needs clearly due to a number of reasons.

Based on our earlier works [19], [20] we noticed that the activities of the user can be used to determine his information needs. These activities show the expectations that the user has in reference to the need he has at hand. Therefore the model we are proposing is a user model that learns from the activities of the user and uses it to improve the existing user model in order to help him with his ISP.

Understanding a problem implies defining the problem in relation to the context and the person who is posing the problem [9].

The problem domain of an EIS consists of:

User, U
A set A of information available in the EIS
A subset R, of A, containing the information that totally or partially satisfy the user's objective.

To compute the set R, the system must have access to all required information. The information about the user can be viewed as:

$$U = \{\text{objective, individual characteristics, context, activities}\}$$

While using an EIS system, the user has an *objective* (finding the information that satisfies his needs). He attempts to reach this objective through a set of interactions (*activities*). His general specificities (e.g. interests etc) are contained in *individual characteristics*. Since understanding a problem implies defining it, in terms of its context and the person who is posing the problem [9], we include the context of the problem in *context*.

The activities of a user are a way of demonstrating his objectives to the system. The user's objective also depends on *individual characteristic* and *context*. In general, an objective is based on the user requirement, and is related to problems revealed by the situational context. How a user creates his objectives from these problems and needs also depends on that user's psychological profile. Given these considerations, a dependent relation can be established from the pair *context, individual characteristics* to *objective* at time *t*.

We use the Dynamic Bayesian Nets (DBNs). This is because DBNs allow the state of the system to be represented as a set of hidden and observed states in terms of state variables, among which there can be complex interdependencies. Graphical models are graphs in which the nodes represent random variables and the arcs represent direct dependencies between these random variables. When the arcs are directed, such graphical models are called *Bayesian Networks*. In such networks, each node is associated with a conditional probability distribution. The graphical structure provides a convenient means of specifying the conditional independencies, and hence represents a compact parameterisation of the model.

For our EIS, given the observed user activities and using a prior graphical model, the objective is to infer the user's goal and ultimately to calculate R. The User's objective is a time-dependent hidden variable (an array of hidden variables) of the model that the system must compute in order to compute *R*. The other two time-dependent hidden variables *individual characteristics* and *context* are also relevant in the determination of *R*, and ultimately will be inferred from the activities sequence of the user. In generic terms, the system observes *activities* and attempts to compute:

$$f : (\{\text{objective, individual characteristics, context, activities}\}, A) \rightarrow R$$

With the above parameters, the set R of the information retrieved, will contain information that is not just based on the content of the information but on its relevance to the user's objective. They can also be used to determine the user's preferred interaction modalities, presentation, etc.

### 4.1 An application to a document base

As stated earlier, we use Bayesian networks to realise this model. The reason is simply because these networks can been used in determining user's needs by the integration of user models as used in a number of other applications. The major problem of our type of models is the stochastic representation of *context, individual characteristics* and *objective*. There are other difficulties during the update of the conditional probability table of each node. We are able to get by these problems by assuming that certain information are inferred by statistical reasoning and the others are explicitly acquired. Our major aim is to define the user's objective as a sum of the objectives inferred by the Bayesian networks and the objectives got explicitly from the user's queries.

We imagine a user that would like to have a list of the journals where at least a member of his research team has published at least once. We determine first, the interaction space of the user. Depending on the type of interface implemented, for instance in a menu based scenario, the user might need to furnish certain information such as his profile information to determine his *individual characteristics*. In a rich dialogue style, the system might infer that the person involved is a new PhD student because during their dialogue, keywords like new, student, PhD, university, research etc. were mentioned by the user. During the same conversation with the system, the user might have used keywords like list of journals, research team, publications, etc. therefore his *context* takes the value (list of journals of publications of research team). Further interactions between the user and the system might throw up: year ≥ 2003; members of the research team etc., the system can now deduce that the user who is a new research student with a research team has an *objective* of looking for a list of journals where members of his research team has published in the last three years.

## Conclusion

With the model discussed here, the issue of adapting an EIS to a user model that is trained from the behaviours (activities) of the user is highlighted. Depending on the type of interface that is developed for the system, the population of each of these parameters can be done. While testing this model on an information base containing documents published by members of our research team, we discovered that there are some attributes that certain users will need to respond to some of their information problems that were not included, though we have not yet implemented this model completely. This helps us to conclude that this model can help in determining the adequacy of the information base to respond the user's needs. The next stage of this work is to completely integrate this model in the information base of the research centre and do a comparative study of the results obtained with the results that would have been obtained if the model was not used.